# The First Differentiable Transfer-Based Algorithm for Discrete MicroLED Repair

Ning-Yuan Lue

Rayleigh Vision Intelligence

Hsinchu, Taiwan

ny.lyu@rvi.com.tw

**Abstract**

Laser-enabled selective transfer allows high-throughput microLED placement, but requires algorithms that can plan shift sequences to minimize motion of the XY stages and adapt to varying optimization objectives across the substrate. We propose the first repair algorithm based on a differentiable transfer module designed to model discrete shifts of transfer platforms, while remaining trainable via gradient-based optimization. Compared to local proximity searching algorithms, our approach achieves superior repair performance and enables more flexible objective designs, such as minimizing the number of steps. Unlike reinforcement learning (RL)-based approaches, our method eliminates the need for handcrafted feature extractors and trains significantly faster, allowing scalability to large arrays. Experiments show a 50% reduction in transfer steps and sub-2-minute planning time on 2000×2000 arrays. This method provides a practical and adaptable solution for accelerating microLED repair in AR/VR and next-generation display fabrication.

## Introduction

Micro-light-emitting diodes (microLEDs or µLEDs) are rapidly emerging as a next-generation display technology due to their high brightness, low power consumption, and excellent scalability across device sizes. In particular, µLEDs are considered a leading candidate for augmented and virtual reality (AR/VR) displays, where pixel-level precision and energy efficiency are crucial. To fabricate such displays, millions of µLEDs must be accurately transferred from donor wafers onto target backplanes—a process known as mass transfer [1, 2].

During mass transfer, process variations and imperfect adhesion often result in placement defects that must be corrected through selective repair. In laser-enabled transfer systems, µLEDs are temporarily held on a transparent carrier glass coated with a dynamic release layer and selectively released by pulsed laser illumination to fall onto designated sites on the panel. The optical beam is rapidly scanned using high-speed galvanometric mirrors (galvos), while donor and target substrates are positioned via precision XY stages capable of micrometer-level alignment. This high-precision mechanical positioning is essential—if the substrate alignment is off by even a few microns, µLEDs will not land at the intended sites, regardless of laser accuracy [3, 4].

While the laser scanning itself can process 2,000–20,000 μLEDs per second, each repositioning of the transfer stage typically incurs a delay of 1 to 5 seconds. As such, stage movement becomes the bottleneck in overall throughput. This motivates the need for intelligent algorithms that plan shift sequences to reduce cumulative stage movement while ensuring complete repair coverage.

Existing approaches often rely on rule-based heuristics that are fast but inflexible—they struggle to adapt to varying optimization goals such as minimizing directional movement cost, reducing shift count, or accounting for asymmetries in platform speed. Recently, reinforcement learning (RL)–based methods have been proposed for similar industrial planning tasks. While they offer flexibility through end-to-end learning, they often suffer from long training times and unstable performance, especially when raw state inputs lack well-structured features. In such settings, additional domain-specific feature engineering or careful design of the input network becomes necessary, which can hinder generalization and practical deployment.

To address these limitations, we propose a new repair planning algorithm based on **differentiable transfer modeling**. Inspired by advances in differentiable simulation and straight-through estimators (STE) [5,6], our method embeds integer-valued spatial shifts in a fully differentiable computation graph. This allows shift sequences to be optimized using standard gradient-based methods, without relying on hand-designed features or long RL training cycles.

Unlike prior methods, our model can directly incorporate coverage penalties into the loss function. Experiments on synthetic μLED panels show that our approach reduces total transfer steps by 50%, achieves better repair quality than local proximity searching baselines, and completes planning on 2000×2000 arrays in under two minutes. This work provides a scalable and flexible framework for optimizing repair-stage motion in high-throughput microLED assembly lines.

**Related works**

Differentiable optimization has gained attention in various domains for enabling gradient-based learning over non-differentiable components. In differentiable physics simulators, such as [7], the simulation of physical environments is made compatible with backpropagation, allowing policy gradients to flow through dynamics models. In the area of differentiable matching, works like [8] use soft-relaxed operations (e.g., Gumbel-Sinkhorn) to approximate discrete permutations, making them amenable to neural training. However, our method requires executing truly discrete platform shifts over multiple steps, where soft approximations could lead to cumulative alignment errors. Instead, we adopt a rounding mechanism with a straight-through estimator (STE) to maintain both differentiability and discrete actuation.

For comparison, differentiable image registration methods (e.g., [9]) aim to align images using spatial transformations optimized via smooth losses. While we similarly use bicubic interpolation to achieve smoothness in the loss landscape, our setting differs in that we apply uniform rigid shifts across binary LED placement maps, and such shifts are repeated multiple times, making precision and discrete executability critical in manufacturing settings.

## Method

We propose a differentiable discrete-shift repair planning algorithm to simulate repair operations of microLED. The system models the iterative movement of XY stages and the laser-induced replacement of defective pixels. The repair planning algorithm is composed of multiple **differentiable transfer modules** (see Figure a). Each transfer module takes as input the COC1 and COC2 arrays, along with a displacement vector representing the relative shift from the origin of COC1 to the origin of COC2. The module simulates transferring matching LEDs from COC1 to COC2, and outputs the updated COC1 and COC2 arrays. The number of stacked transfer modules corresponds to the number of physical platform alignments performed during the repair process.

During training, the displacement vector of each module is learnable and can be iteratively adjusted to optimize a given objective. However, since these displacements correspond to physical movements of mechanical platforms, the shifts executed must be discrete, integer-valued displacements (e.g., in micrometer units). We enforce discrete-valued behavior using a **rounding function in the forward pass**, while maintaining differentiability in the backward pass via the **Straight-Through Estimator (STE)**.

We first formulate the problem, describe the architecture of the transfer module, then explain how we ensure that the actual displacement remains discrete while retaining differentiability via a Straight-Through Estimator (STE). Finally, we detail the optimization objectives that guide the training process.

### A. Problem Formulation

Let $C_1$ denote the donor chip(chip on carrier 1 or COC1)2D array, and $C_2$ denote the target chip(COC2) array. Each pixel in the arrays takes a binary value: 1 represents the presence of an LED chip at that position, and 0 indicates that the site is empty:

$C_{1,i,j} \in \{0,1\} \text{ if } 0 \leq i \leq W_1 - 1, \text{ and } 0 \leq j \leq H_1 - 1, \quad C_{1,i,j} = 0 \text{ elsewhere,}$

$C_{2,i,j} \in \{0,1\} \text{ if } 0 \leq i \leq W_2 - 1, \text{ and } 0 \leq j \leq H_2 - 1, \quad C_{2,i,j} = 1 \text{ elsewhere,}$

$(i,j) \in \mathbf{Z}^2$

For $C_1$ with the origin initially corresponds to the origin of $C_2$, a shift vector $\mathbf{s} = (s_x, s_y) \in \mathbf{Z}^2$ that moves $C_1$ will make $C_{1,i-s_y,j-s_x}$ corresponding to $C_{2,i,j}$, then the repair action is represented by swap $C_{1,i-s_y,j-s_x}$ and $C_{2,i,j}$ values:

$Reset\ C_{1,i-s_y,j-s_x} = 0,\ C_{2,i,j} = 1\ if\ C_{1,i-s_y,j-s_x} = 1,\ C_{2,i,j} = 0$

Our goal is to find a sequence of 2D shifts $\{s^{(t)}, 1 \leq t \leq T\}$, such that as many defective positions on COC2 are repaired by matching corresponding positions in COC1.

**B. Differentiable transfer Module**

**B.1 Shift the COC1 array with a vector $a$**

Construct the bicubic interpolation of $C_1$: $B_1(x,y) = $ bicubic interpolation of $C_1$, $(x,y) \in R^2$

Then shift $B_1$ with a 2D continuous vector $a$: $D_1(x,y) = B_1(x - a_x, y - a_y)$, $a = (a_x, a_y) \in R^2$.

We take values on integer-valued positions: $D_{1,i,j} = D_1(i,j)$, $(i,j) \in Z^2$

The above operations are denoted as: $D_1 = f_{shift}(C_1; a)$

**B.2 Transfer LEDs from COC1 to COC2**

$D'_{1,i,j} = D_{1,i,j} - D_{1,i,j} \times (1 - C_{2,i,j})$

$C'_{2,i,j} = C_{2,i,j} + D_{1,i,j} \times (1 - C_{2,i,j})$,  $(i,j) \in Z^2$

The transferred array $D'_1$ and $C'_2$ are then obtained. We denote the operations as:

$(D'_1, C'_2) = f_{transfer}(D_1, C_2)$

**B.3 Shift the COC1 array back with the vector $-a$**

Construct the bicubic interpolation of $D'_1$: $B'_1(x,y) = $ bicubic interpolation of $D'_1$, $(x,y) \in R^2$

Shift the $B'_1$ with the vector $-a$, and take values at integer-valued positions to form $C'_1$:

$C'_{1,i,j} = B'_1(i + a_x, j + a_y)$, $(i,j) \in Z^2$. These operations are denoted as:

$C'_1 = f_{shift}(D'_1; -a)$

**B.4 Discretize the shift action with STE**

Given a 2D continuous shift vector $v = (v_x, v_y) \in R^2$. To make the shift vector input to transfer

module integer-valued, we round it to the nearest integer:

$a_x = f_{round}(v_x), \ a_y = f_{round}(v_y)$, such that $\boldsymbol{a} = (a_x, a_y) \in \boldsymbol{Z}^2$.

However, $f_{round}$ is not differentiable, so we use the STE trick, just define $\frac{df_{round}(v)}{dv} \equiv 1$ when doing back propagation.

Input $\boldsymbol{C}_1, \boldsymbol{C}_2$ and $\boldsymbol{v} = (v_x, v_y)$, the complete differentiable transfer module is:

$\boldsymbol{a} = (f_{round}(v_x), f_{round}(v_y))$

$\boldsymbol{D}_1 = f_{shift}(\boldsymbol{C}_1; \boldsymbol{a})$

$(\boldsymbol{D'}_1, \boldsymbol{C'}_2) = f_{transfer}(\boldsymbol{D}_1, \boldsymbol{C}_2)$

$\boldsymbol{C'}_1 = f_{shift}(\boldsymbol{D'}_1; -\boldsymbol{a})$

We will simply denote these as $(\boldsymbol{C'}_1, \boldsymbol{C'}_2) = F_{transfer}(\boldsymbol{C}_1, \boldsymbol{C}_2; \boldsymbol{v})$. Notice that $\boldsymbol{v}$ is a learnable vector to optimize the target.

## C. Differentiable repair planning model and training objective

The repair algorithm applies the transfer module repeatedly to the input COC1 and COC2 arrays. A repair model performing T transfer steps can be expressed as:

$\boldsymbol{C}_1^{(1)} = \boldsymbol{C}_1, \ \boldsymbol{C}_2^{(1)} = \boldsymbol{C}_2,$

$(\boldsymbol{C}_1^{(t+1)}, \boldsymbol{C}_2^{(t+1)}) = F_{transfer}(\boldsymbol{C}_1^{(t)}, \boldsymbol{C}_2^{(t)}; \boldsymbol{v}^{(t)}) \quad , \quad 1 \leq t \leq T$

The optimization objective is to maximize the number of LEDs on the COC2 array. To encourage the repair to complete within $T_1 < T$ transfer steps, the loss function can be formulated as:

$L_1(\boldsymbol{C}_1, \boldsymbol{C}_2; \{\boldsymbol{v}^{(t)}\}) = -\sum_{\substack{0 \leq i \leq W_2-1 \\ 0 \leq j \leq H_2-1}} [C_{2,i,j}^{(T_1+1)} + \lambda_1(C_{2,i,j}^{(T+1)} - C_{2,i,j}^{(T_1+1)})], \ \lambda_1 < 1.$

The optimization objective can be flexibly designed. For example, if the goal is to reduce platform displacement, an additional penalty term can be incorporated as:

$L_2(\{\boldsymbol{v}^{(t)}\}) = \lambda_2 \sum_{t=2}^{T} ||\boldsymbol{v}^{(t)} - \boldsymbol{v}^{(t-1)}||_1$ , where $\lambda_1$ and $\lambda_2$ are hyperparameters.

The optimized shift sequence is $\{\boldsymbol{s}^{*(t)}, 1 \leq t \leq T\}$,

$s_x^{*(t)} = f_{round}(v_x^{*(t)}), \ s_y^{*(t)} = f_{round}(v_y^{*(t)}), \ 1 \leq t \leq T$

$$\{v^{*(t)}\} = \operatorname*{argmin}_{\{v^{(t)}\}}(L_1 + L_2)$$

The training procedure and stopping criteria are detailed in the experimental setup (Section A.1).

## D. Baseline Algorithms for Comparison

### D.1 Local proximity Searching

As a baseline, we implement a simple algorithm that iteratively finds and executes individual transfer actions based on local proximity. At each step, we select an unfilled target site $x_2$ on the COC2 array, and identify the nearest available source chip at $x_1$ on the COC1 array that minimizes the Euclidean distance between $x_1$ and $x_2$. The corresponding shift vector is computed such that the chip at position $x_2$ is transferred to position $x_1$. After the transfer, the COC1 and COC2 arrays are updated accordingly. This method does not consider the global correlation between arrays and optimizes each step without planning future actions. While simple and fast to execute, it may lead to suboptimal repair sequences, especially when multiple candidate matches exist or when minimizing total movement cost is critical.

Get one of the minimums of $C_2$ (empty site on COC2): $x_2 = (i^*, j^*) = \operatorname*{argmin}_{(i,j)} C_{2,i,j}$.

Find the position on $C_1$ with a chip on which is nearest to $(i^*, j^*)$:

$$x_1 = (k^*, l^*) = \operatorname*{argmin}_{(k,l)}[(k - i^*)^2 + (l - j^*)^2] \text{ with } C_{1,k,l} = 1.$$

The shift with which to execute transfer action is $(s_x, s_y) = x_2 - x_1$

### D.2 Reinforcement Learning -Based Method

To provide a learning-based baseline for comparison, we implemented a reinforcement learning (RL) agent that learns to plan transfer shifts through interaction. We adopt Proximal Policy Optimization (PPO) [11], a widely used policy gradient method known for its sample efficiency and stable performance in high-dimensional action spaces. PPO balances exploration and exploitation by limiting the policy update at each iteration, preventing drastic changes that can destabilize training.

At each step, the agent observes the current COC1 and COC2 binary arrays and selects a shift action for the transfer platform. The objective is to minimize the total number of transfer steps while ensuring effective repair coverage.

The agent's policy is represented by a convolutional neural network (CNN) [11] that encodes the state into features and outputs shift directions as discrete actions. Specifically, for a 50×50 array, the network takes two binary maps (COC1 and COC2) as input channels and produces a categorical distribution over a 99×99 grid of possible integer shifts (e.g., from (−49, −49) to (+49, +49)). Actions are sampled from this distribution during training.

We design the reward function to encourage progress in COC2 coverage. For each step, the agent receives a positive reward proportional to the number of newly repaired pixels in COC2, and a small penalty for each step to encourage efficiency. A final reward is given when full coverage is achieved.

While this approach offers flexibility and learns repair heuristics automatically, it requires many episodes to converge and can be sensitive to hyperparameters. Compared to our differentiable model, the RL baseline exhibits significantly longer training time and less stability across runs.

**Experiments**

The optimization objective is first set to **maximize the final number of LED chips on COC2**. Under this objective, we evaluate how many transfer steps are required by different repair algorithms given COC1 and COC2 arrays of varying sizes and defect rates. Then **the loss landscape** is **visualized** with respect to the learnable shift parameters and provide a qualitative analysis of convergence behavior under discrete action approximation.

**A. Repair Efficiency Across Array Sizes and Defect Rates**

We vary the sizes and defect rates of the COC1 and COC2 arrays and compare the number of transfer steps required by three methods: a local proximity Searching, reinforcement learning (RL), and our differentiable transfer (DT) model. Let the dimensions of the source and target arrays be denoted by $(W_1, H_1)$ and $(W_2, H_2)$, respectively. The initial defect rates of COC1 and COC2 are defined as:

COC1 defect rate $d_1 = \frac{1}{W_1 H_1} \sum_{\substack{0 \leq i \leq W_1-1, \\ 0 \leq j \leq H_1-1}} (1 - C_{1,i,j})$

COC2 defect rate = $d_2 = \frac{1}{W_2 H_2} \sum_{\substack{0 \leq i \leq W_2-1, \\ 0 \leq j \leq H_2-1}} (1 - C_{2,i,j})$

**A.1 Differentiable repair planning (DRP)**

In the training phase, we begin by randomly generating a pair of COC1 and COC2 arrays according to the specified array size and defect rates. Starting with T = 3 differentiable transfer modules, we construct a repair model consisting of T shift operations. Each learnable shift parameter $\{v^{(t)}, 1 \leq t \leq T\}$ is initialized to zero. The model is trained using Adam optimizer for 300 iterations to minimize a specified loss objective (e.g., total number of defects remaining on COC2 after T transfers). After training, we record the minimum loss value achieved and the corresponding shift sequence. If the loss has not yet converged (e.g., nonzero residual defects), we increase T and repeat the process. This iterative procedure continues until the minimum loss converges, which indicates the smallest number of transfer steps required to complete the repair task. The whole process is listed in Algorithm 1.

**Algorithm 1: Adaptive Training of Differentiable Repair Model**

**Input:**

- Initial defect rates $d_1$ and $d_2$
- COC1 and COC2 array sizes $(W_1, H_1)$ and $(W_2, H_2)$
- Maximum allowed transfer steps $T_{max} = 100$
- Maximum optimization iterations per T: $N_{iter} = 1000$
- Convergence threshold $\varepsilon = 10^{-6}$

**Output:**

- Minimum number of shifts $T^*$
- Optimized shift vectors $\{v^{*(t)}, 1 \leq t \leq T^*\}$

**Steps:**

Initialize $T \leftarrow 3$

**Repeat** until $T > T_{max}$:

Initialize shift parameters $v^{(t)} \leftarrow 0, \ 1 \leq t \leq T$

**For** $i = 1 \ to \ N_{iter}$:
a. Apply $T$ transfer modules with current $\{v^{(t)}\}$ on input COC1, COC2
b. Compute loss (e.g., number of defects on COC2)
c. Use gradient-based optimization (e.g., Adam) to update $\{v^{(t)}\}$

Record the lowest loss $L_{min}$ across iterations

**If** $L_{min} < \varepsilon$:

Return current $T$ and corresponding $\{v^{(t)}\}$

Increment $T \leftarrow T + 1$

For each configuration of array size and defect rate, we generate 100 randomized COC1 and COC2 pairs. The average and standard deviation of the required transfer steps are reported.

### A.2 Local proximity searching (LPS)

Both COC1 and COC2 are initialized with uniformly random defects based on the specified defect rates. The local proximity searching baseline follows the procedure described in Section D.1 of the Methods, sequentially applying the best shift at each step to maximize immediate defect removal on COC2. The repair process continues until all defects in COC2 are resolved.

To evaluate its performance, we apply the local proximity searching strategy on the same set of 100 randomized COC1 and COC2 instances used in the differentiable model experiments. The number of required repair steps is collected across all trials, and the mean and standard deviation are reported for comparison.

### A.3 Reinforcement learning (RL)

We implement a reinforcement learning (RL) baseline using the open-source **Stable-Baselines3 (SB3)** framework. A custom environment is developed to simulate the repair process, where the state consists of the current COC1 and COC2 arrays, and the agent selects a shift vector at each step. A custom CNN-based feature extractor is used to encode the spatial structure of the binary arrays. The feature extractor consists of:

- A **shared CNN** encoder with 3 convolutional layers (kernel size 3, stride 1, ReLU activation), followed by a flattening layer
- Separate **actor** and **critic** MLP heads with 2 fully connected layers (size 128 and 64)

We train a PPO agent on an **NVIDIA RTX 4080 GPU** using SB3 with default hyperparameters (learning rate 3e-4, batch size 2048, etc.). Full settings are provided in the appendix. Training data is collected via online interaction between the agent and environment, using on-policy updates. The reward function is defined as -1 per time step until the repair task is completed, encouraging the agent to minimize the number of transfer steps. No additional terminal reward is provided.

During inference, we evaluate the trained policy on the same 100 randomized COC1/COC2 array pairs used in the differentiable model experiments (Section A.1). We record the mean and standard deviation of the number of transfer steps required to complete the repair for each method.

Table 1 Hyperparameters for RL training

| Parameter | Value |
|---|---|
| Learning rate | 3e-4 |
| Discount factor (γ) | 0.99 |
| GAE λ | 0.95 |
| Batch size | 2048 |
| Epochs per update | 10 |
| Clip range | 0.2 |
| Entropy coefficient | 0.01 |
| Total timesteps | 200,000 |

**A.4 Results**

In all experiments, $W_1 = H_1 = W_2 = H_2$.

| Condition | DRP(ours) | LPS | RL |
|---|---|---|---|
| $W_1 = 50$, $d_1 = 0.1, d_2 = 0.05$ | 3.0 | 3.0 | 3.4 |
| $W_1 = 50$, $d_1 = 0.35, d_2 = 0.05$ | 5.0 | 5.9 | 7.4 |
| $W_1 = 50$, $d_1 = 0.6, d_2 = 0.05$ | 7.1 | 10.6 | 16.2 |
| $W_1 = 100$, $d_1 = 0.1, d_2 = 0.05$ | 3.0 | 3.6 | 3.0 |
| $W_1 = 100$, $d_1 = 0.35, d_2 = 0.05$ | 5.1 | 7.1 | 9.3 |
| $W_1 = 100$, $d_1 = 0.6, d_2 = 0.05$ | 11.1 | 14.4 | 18.4 |
| $W_1 = 500$, $d_1 = 0.1, d_2 = 0.05$ | 4.0 | 5.4 | 5.3 |
| $W_1 = 500$, $d_1 = 0.35, d_2 = 0.05$ | 8.1 | 10.9 | 12.3 |
| $W_1 = 500$, $d_1 = 0.6, d_2 = 0.05$ | 18.0 | 22.8 | 25.1 |

Table 2　Average number of repair steps

**Key observations are as follows:**

**DRP consistently outperforms both baselines, with widening advantages in more complex scenarios.**

While all methods perform comparably in low-defect, small-array regimes (e.g., 100×100 array size, 10% defect density), DRP increasingly surpasses LPS and RL as array size and defect density grow. For instance, at 100×100 with 30% defects, DRP requires 7.1 steps versus 10.6 (LPS) and 16.2 (RL); at 500×500 with 30%, the gap further widens to 18.0 versus 22.8 and 25.1, respectively. This trend demonstrates DRP's ability to globally optimize shift sequences under more constrained or large-scale conditions.

**RL exhibits unstable performance and poor scalability, while LPS suffers from short-sighted heuristics.**

Although RL performs competitively in isolated settings (e.g., 200×200, array size, 10% defect density), it generally degrades with increasing problem complexity due to sparse reward signals and sensitivity to feature extractor design. LPS, while simple and fast, lacks foresight and accumulates redundant movements, leading to suboptimal shift sequences—especially in high-density or large-array repairs.

**DRP scales robustly and remains practical in real-time applications.**

The DRP model optimizes shifts via gradient descent per instance, without the need for pre-training. Although its "inference" entails per-sample optimization, convergence is typically achieved within two minutes for arrays of size 1000×1000. This makes it feasible for dynamic repair tasks, with added flexibility to incorporate application-specific objectives such as asymmetric movement cost.

Compared to reinforcement learning methods, our differentiable approach directly optimizes a global objective with dense gradient signals, enabling faster and more stable convergence. In contrast, RL agents rely on delayed and sparse rewards, often requiring thousands of interaction steps to learn transferable policies. The explicit, differentiable nature of our formulation ensures better sample efficiency, generalization, and controllability—making it well-suited for deployment in real-world microLED repair systems.

### B. Loss Landscape Visualization and Discrete Gradient Effects

Understanding why differentiable shift parameters trained via a Straight-Through Estimator (STE) converge effectively—despite the underlying loss being defined over discrete actions—is a nontrivial theoretical and empirical question [5,6]. Prior work in deep learning suggests that convergence often depends on the presence of locally flat regions in the loss landscape, where many parameter combinations yield similarly low loss values and gradient descent is not strongly impeded by sharp barriers [12-14]. To investigate whether our differentiable transfer (DT) model exhibits such properties, we conduct a series of visualizations and measurements.

### B.1 Global Loss Landscape

We first visualize the global loss landscape for different layers by sweeping the two-dimensional shift parameter (sx,sy) over a wide range and evaluating the corresponding loss. The resulting heatmaps (Fig.3) show broad valleys with low loss values, suggesting that the trained model converges to a globally favorable region.

**B.1 Global Loss Landscape**

We first visualize the global loss landscape for each transfer block by sweeping the two-dimensional shift parameter over the full range of integer shifts ($-49 \leq s_x \leq 49$, $s_y \leq 49$, $-49 \leq s_y \leq 49$) and evaluating the corresponding loss. The resulting heatmaps (Fig. 3) reveal broad valleys with low loss values, indicating that the optimizer operates in a landscape where large contiguous regions correspond to near-optimal outcomes. This structure likely facilitates navigation toward solutions close to the global minimum.

**B.2 Local Landscape and Trajectory Visualization**

We next zoom into the local neighborhood around each optimized shift vector and plot high-resolution loss landscapes (Fig. 4). These plots are overlaid with the actual optimization trajectories over the first 100 epochs, starting from the initial position (0,0) (green dot), moving along the continuous-valued updates (yellow curve), and ending at the final learned shift (purple dot), with the corresponding rounded integer shift for execution marked in red.

In this repair problem, the loss is defined as the number of unrepaired LEDs remaining in COC2. Since the value of a perfect or near-perfect repair is known a priori (i.e., zero or close to zero), we can set a threshold on the loss and terminate optimization once it falls below this value. Consequently, the process can be viewed less as unconstrained gradient descent toward an unknown optimum and more as a gradient-guided search: the optimizer gradually moves into a low-loss region and continues exploring until it finds a point below the known loss threshold.

**B.3 Gradient Convergence Behavior**

To analyze convergence more directly, we track the shift parameters across training epochs. Figure 5 plots the Euclidean norm of each block's shift vector as a function of training epoch. For all blocks, the shift magnitude stabilizes after a certain number of iterations, indicating that the STE-based gradient estimates are sufficient to guide the parameters into a stable region.

Interestingly, even without early stopping, the gradient magnitude decreases from an initial scale of $O(1)$ to below 1e-4 after roughly 100 epochs, suggesting that the algorithm remains stationary and does not wander once it has entered a low-loss region. While a full theoretical account is beyond the scope of this work, this behavior is likely due to the use of a **surrogate loss** and **surrogate gradients**—computed from a bicubic-interpolated landscape near integer-valued shifts—which effectively smooth the discrete loss surface. How such surrogate gradients influence convergence in multi-step discrete planning remains an open question for future study.

**Conclusion**

This work introduces the first differentiable repair algorithm for microLED platform-based selective transfer. By modeling each discrete alignment step as a differentiable shift module, the proposed method enables end-to-end optimization of repair trajectories using gradient-based training. Unlike conventional methods, it supports flexible design goals and operates efficiently on large arrays, achieving significant reductions in repair steps and planning time.

Compared to reinforcement learning-based approaches, our method avoids the need for explicit feature extraction networks and extensive sampling strategies, leading to faster training and improved scalability. Although inference time is slightly higher, the differentiability allows better convergence behavior and generalization. Additionally, our approach can serve as a strong initialization or supervised target for RL-based methods. In contrast to rule-based algorithms, it allows direct incorporation of physical constraints and objective trade-offs, such as asymmetric movement costs.

This framework opens new possibilities for high-precision repair in AR/VR and display manufacturing. Future work may explore its integration into hybrid planning systems, online adaptation to real hardware feedback, or extensions to non-uniform array layouts and probabilistic defect modeling.

Figure 1: Laser-enabled transfer system

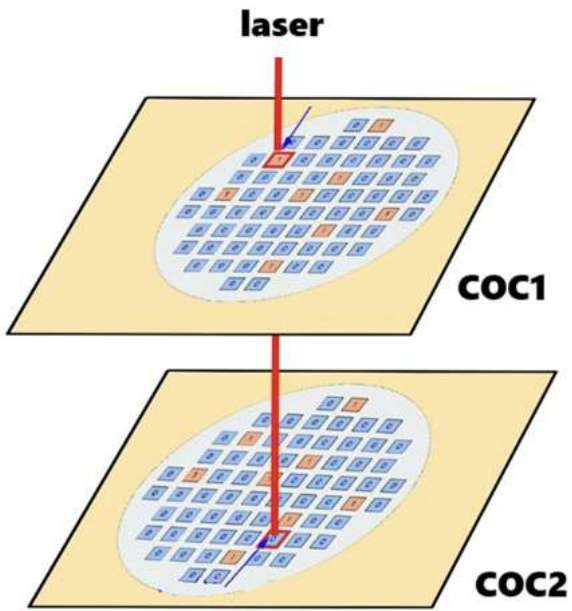

Schematic diagram of the laser-enabled selective transfer system for microLED repair. The donor chip-on-carrier (COC1) is positioned above the target substrate (COC2), with a pulsed laser incident from the top. The laser beam is aligned such that its extended path passes through corresponding sites on both COC1 and COC2, which are highlighted with blue arrows to indicate the transfer target.

Figure 2: Differentiable transfer module and repair planning model

(a)

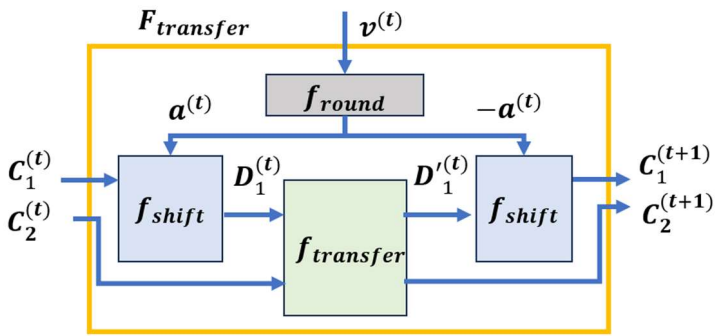

(b)

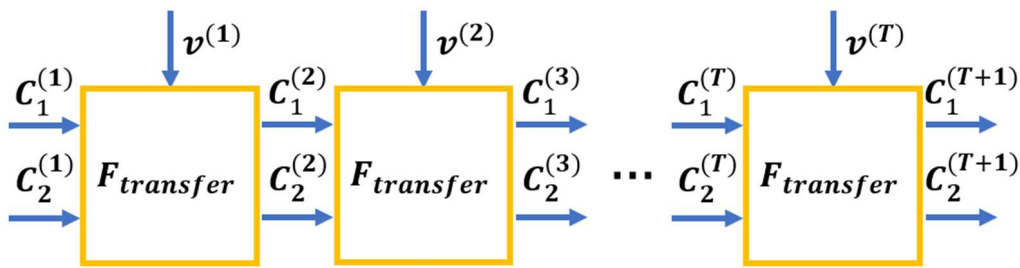

(a) Structure of a differentiable transfer module, which takes the current COC1 and COC2 arrays along with a learnable 2D shift vector, applies bicubic interpolation for continuous shifting, transfers matching LEDs from COC1 to COC2, and enforces discrete actuation via a Straight-Through Estimator. (b) Differentiable repair planning model composed of multiple transfer modules, each corresponding to one physical alignment step in the repair sequence.

Figure 3: Global loss landscape

(a)

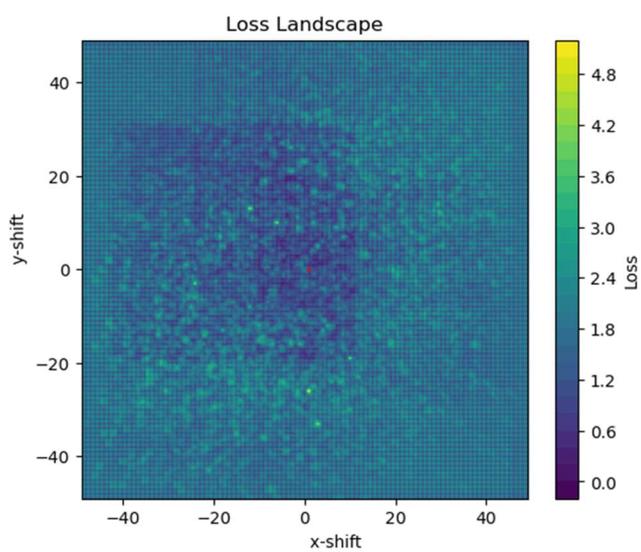

(b)

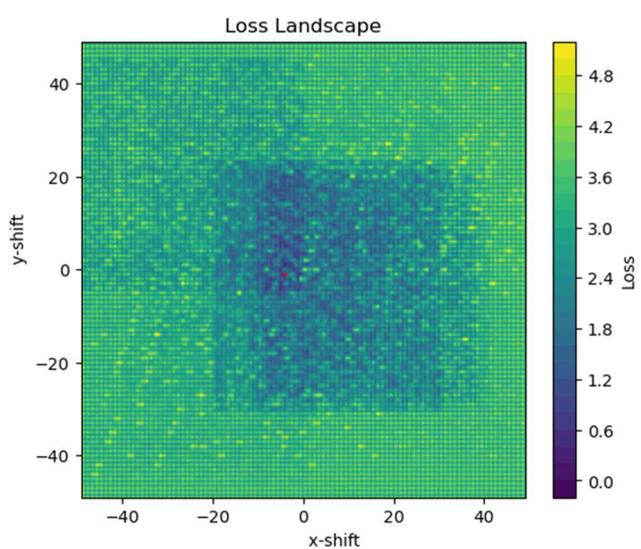

(c)

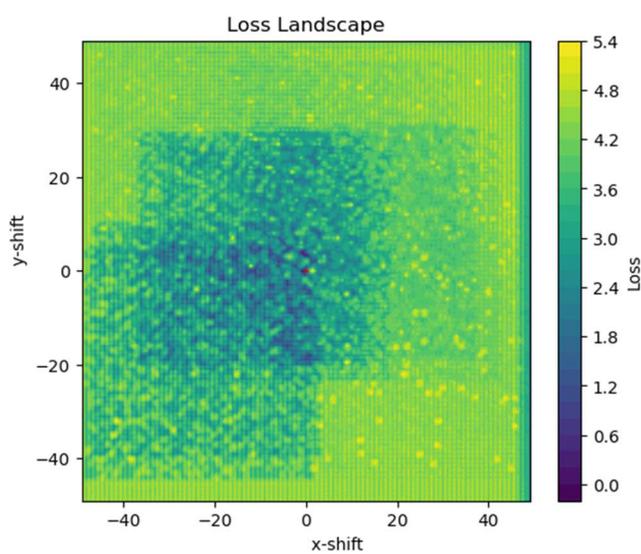

(d)

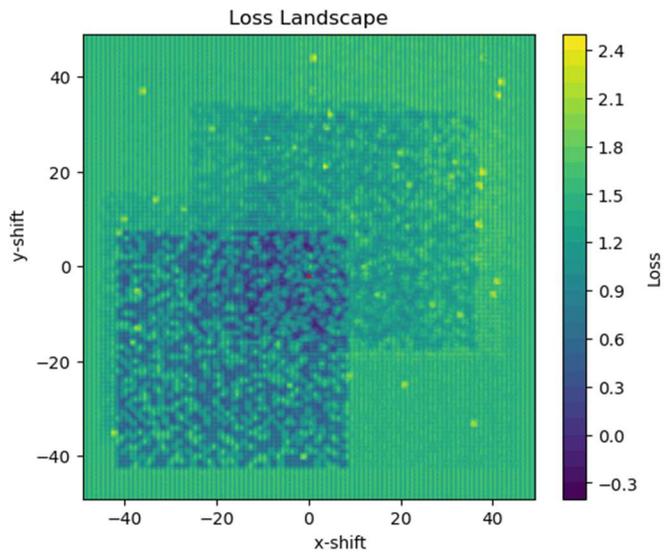

(e)

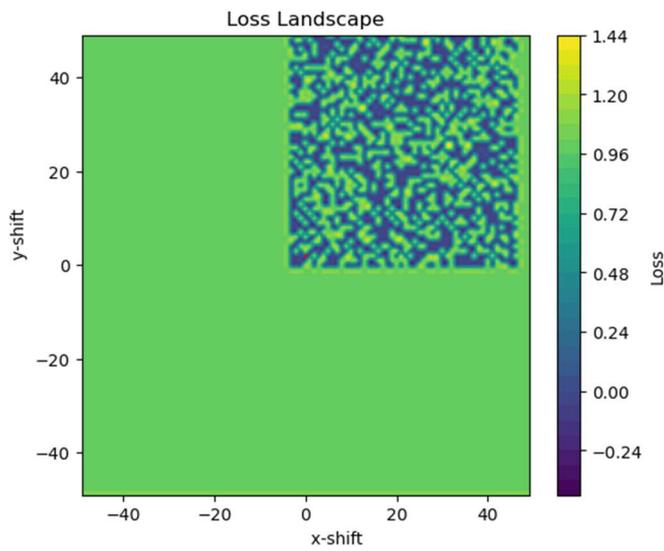

Global loss landscapes for each of the five transfer blocks in a 50×50 repair task (T=5). For each block, the loss is plotted over the 2D space of x-shift and y-shift values, with color indicating the loss magnitude (darker colors represent lower loss). The red dot marks the optimized shift vector obtained after training. Subfigures (a)–(e) correspond to the first through fifth transfer blocks, respectively, showing how the optimization target evolves across successive repair steps.

Figure 4: Local Landscape and Trajectory

(a)

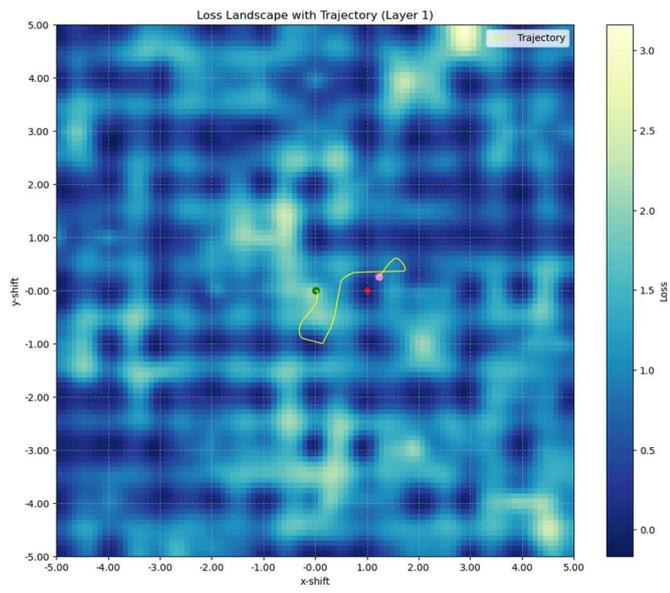

(b)

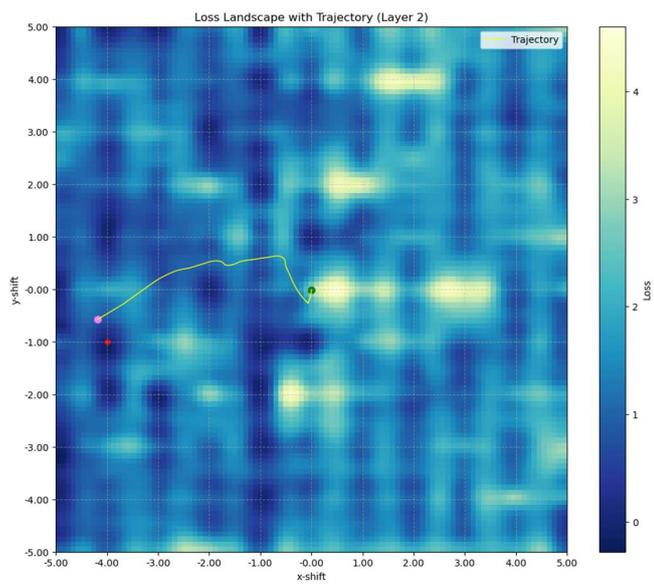

(c)

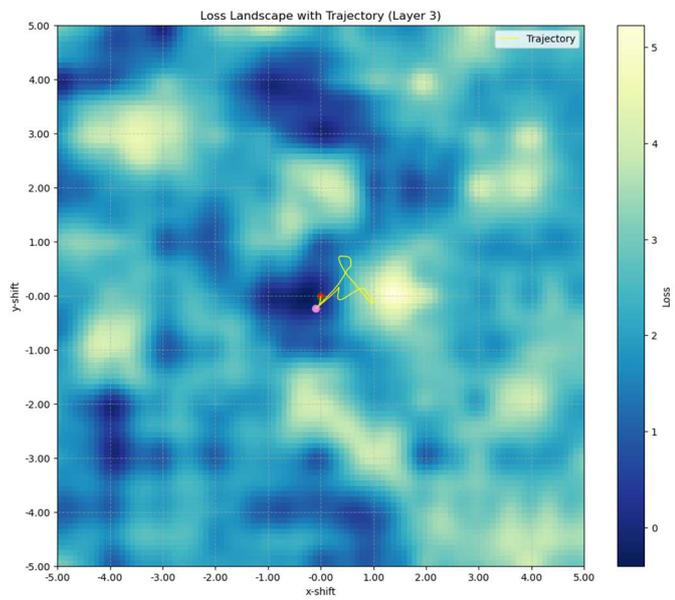

(d)

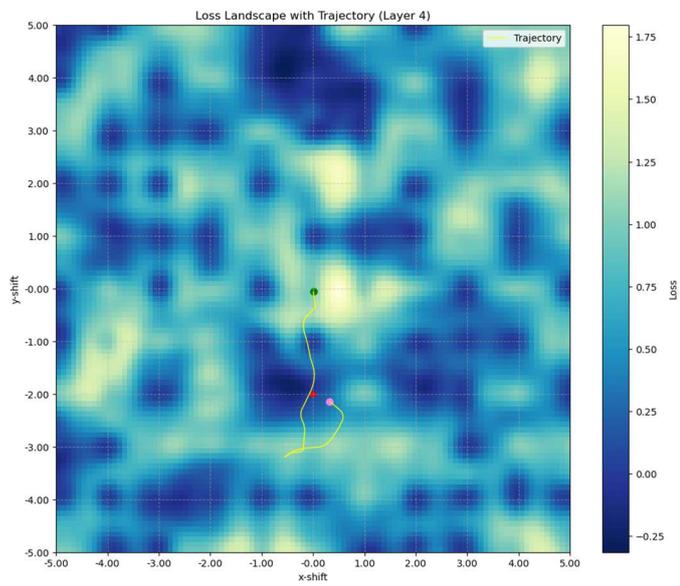

(e)

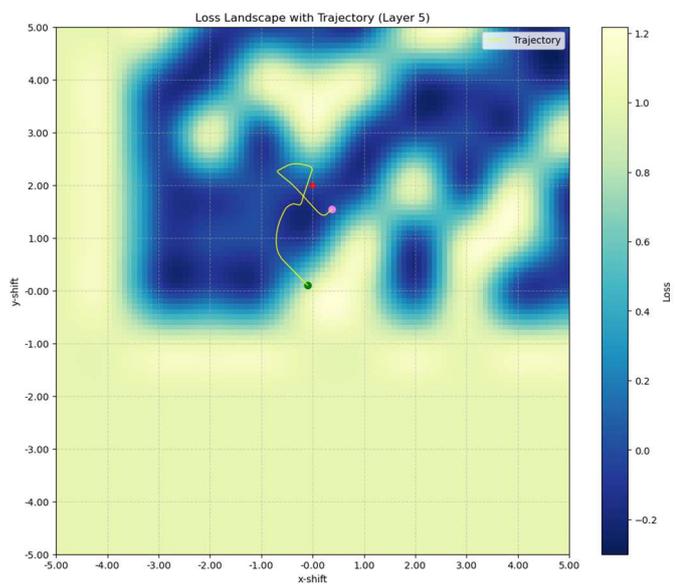

High-resolution, zoomed-in loss landscapes for the five transfer blocks in the 50×50 repair task, corresponding to Figure 3(a)–(e). Each plot focuses on a local region around the optimum in the x-shift/y-shift plane. Color shading denotes loss magnitude, with darker regions indicating lower loss. The yellow curve traces the evolution of the shift vector over the first 100 training epochs, starting from the green dot at the initial position (0,0) and ending at the purple dot representing the final continuous-valued shift. The red dot marks the rounded integer shift used for physical execution.

Figure 5: Convergence of the gradient

(a)

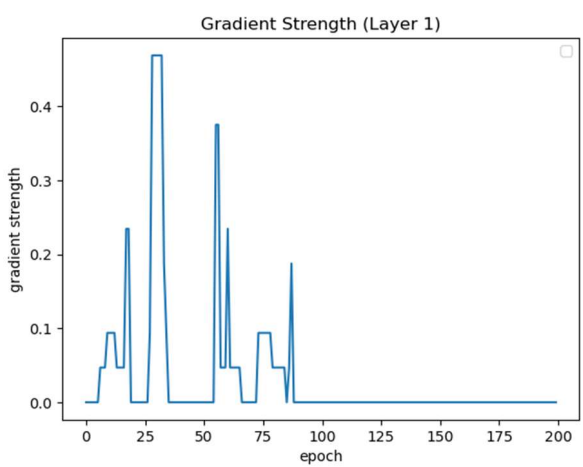

(b)

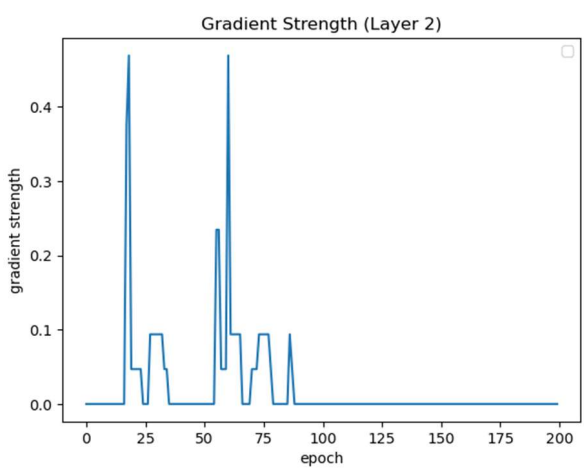

(c)

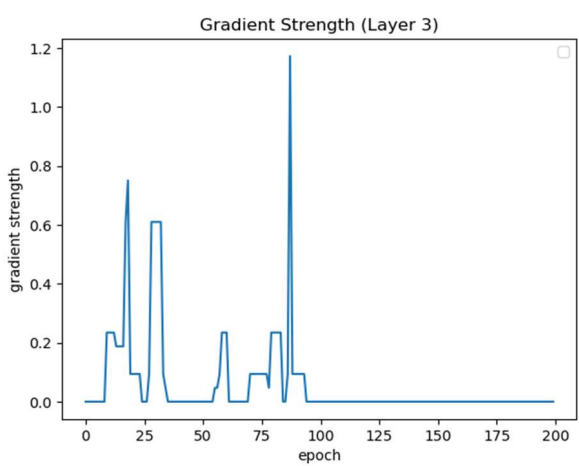

(d)

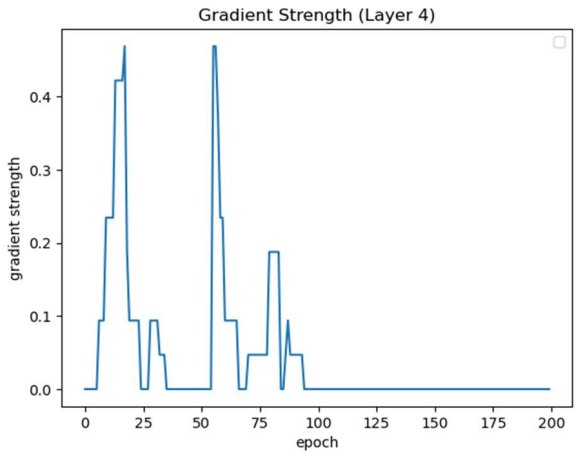

(e)

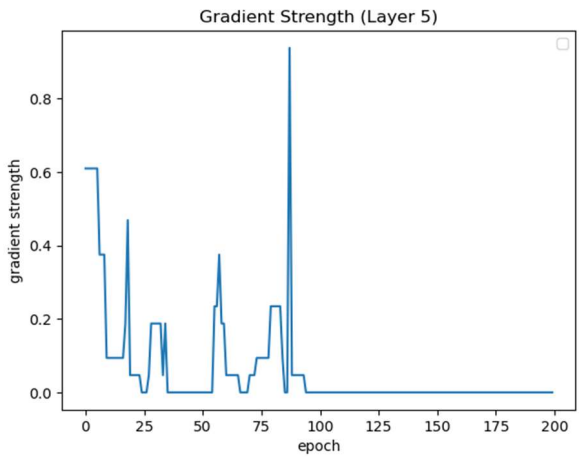

Shift vector magnitude as a function of training epochs for each transfer block in the 50×50 repair task. Subfigures (a)–(e) correspond to the first through fifth transfer blocks, respectively. The horizontal axis represents the training epoch, and the vertical axis shows the Euclidean norm of the shift vector each iteration. The plots illustrate how the learned shifts converge over time, indicating stable optimization behavior across all blocks.